# TWO-FACTOR AUTHENTICATION SMART ENTRYWAY USING MODIFIED LBPH ALGORITHM

Zakiah Ayop [1], Wan Mohamad Hariz Bin Wan Mohamad Rosdi [2],
Looi Wei Hua [3], Syarulnaziah Anawar [1], Nur Fadzilah Othman [1]

[1] Fakulti Teknologi Maklumat dan Komunikasi, Universiti Teknikal Malaysia Melaka, Melaka, Malaysia
[2] ASK Pentest Sdn. Bhd., Kuala Lumpur, Malaysia
[3] V6 Technology Sdn Bhd, Penang, Malaysia

## ABSTRACT

*Face mask detection has become increasingly important recently, particularly during the COVID-19 pandemic. Many face detection models have been developed in smart entryways using IoT. However, there is a lack of IoT development on face mask detection. This paper proposes a two-factor authentication system for smart entryway access control using facial recognition and passcode verification and an automation process to alert the owner and activate the surveillance system when a stranger is detected and controls the system remotely via Telegram on a Raspberry Pi platform. The system employs the Local Binary Patterns Histograms for the full face recognition algorithm and modified LBPH algorithm for occluded face detection. On average, the system achieved an Accuracy of approximately 70%, a Precision of approximately 80%, and a Recall of approximately 83.26% across all tested users. The results indicate that the system is capable of conducting face recognition and mask detection, automating the operation of the remote control to register users, locking or unlocking the door, and notifying the owner. The sample participants highly accept it for future use in the user acceptance test.*

## KEYWORDS

*Facial Recognition, Face Mask Detection, Access Control, Internet of Things, Raspberry Pi, Smart Door Access Control*

## 1. INTRODUCTION

According to the Department of Statistics Malaysia, there were 40,465 property crime cases in Malaysia in 2022. Property crimes include house break-ins and theft, vehicle theft, and other theft. However, 10,585 cases out of the overall number of property crimes comprise house break-ins and theft [1]. It represents 26.15% of Malaysia's total property crime in 2022.

Currently, most residents have a door lock containing only a one-factor authentication method, which is insufficient to ensure the residential security and protection of household members. Besides, some entryway access control systems are only activated when the doorbell is pressed. The existing surveillance system may be compromised by burglary, and the property owner may not receive timely notification of unauthorized entry. Lastly, there is a lack of remote control in the entryway access control system. The owner cannot remotely control the system via the mobile application when not at home.

Smart entryways are essential because they can keep up with the changing security needs of homes. With burglaries and unauthorised access becoming more of a worry, the system offers more protection than traditional locks. Biometric identification, such as face recognition, fingerprints, or eyes, adds an extra layer of security by ensuring that only registered people are allowed to enter [2]. Many studies in IoT proposed 2FA systems using face recognition and other factors, such as passcodes [3][4][5] or RFID [4][5]. However, there is a lack of implementation in face mask detection in this area [3][6][7]. The difficulties lie at the image dataset does not have complete facial information [6].

The main contributions are as follows:

i) design and implement a two-factor authentication system for smart entryway access control using facial recognition for full and occluded face and passcode verification. The full face recognition use in OpenCV LBPH algorithm while the occluded face recognition will use a modified LBPH algorithm proposed in this system. With the modified LBPH algorithm, it can detect occluded faces with good accuracy.

ii) develop an automation process to alert the owner and activate the surveillance system when a stranger is detected. This is enabled by detecting the intruder using PIR sensor and alert the owner through the Telegram.

iii) enable remote control of the system via Telegram on a Raspberry Pi platform. The user management such as add new user, training face dataset, set and reset password and delete user is enabled in this module.

This paper is organized as follows. Section 2 discusses the research background in Two-Factor Authentication Smart Entryway and Face Recognition algorithms. Section 3 introduces the proposed system overview. Section 4 outlines the discussion of testing and results, and section 5 concludes the overall of this study.

## 2. RELATED WORK

An entryway access control system, or, in conventional terms, door lock access control, is a system that is used to restrict entrance to and from a restricted location. Most conventional entryway access control systems rely on a single authentication factor, known as Single-Factor Authentication, which is a lock-and-key system, where the door can be unlocked by any individual possessing the correct physical key. This system has security vulnerabilities, as lost or stolen keys can be copied or given to unauthorised individuals. To address this problem, it is necessary to re-key the locks when a mechanical key is lost. SFA is regarded as a low-security authentication method and should be avoided [8]. In comparison to single-factor authentication, which relies solely on a single authentication approach, two-factor authentication provides a superior level of security by employing two distinct authenticating factors [9][10].

With the advancement of the Internet of Things (IoT), a digital entryway access control system can be developed to lock and open doors without using a physical key. The system have several features that boost security, such as RFID, biometrics, and IoT sensors. Additionally, it can allow only authorised users to access the system and prevent unauthorised access [11].

Several features have been implemented in the previous work, including motion detection, face recognition, two-factor authentication (2FA), remote lock, and notification.

## 2.1. Motion Detection

Motion sensors are popular in terms of security and energy efficiency. When a PIR motion sensor detects motion in the environment, it can activate burglar alarms or security cameras. It can save energy by hibernating the system when it detects no movement.

An IoT-based security alert system has been proposed using an intrusion detection system approach [12][13]. The system detects motion inside a space and alerts the owner if an intruder exists. The motion sensor triggers the connected camera to record and save images in the system's temporary storage. Another research [14] proposed a system in which when the PIR sensor detects the presence of a human at the entrance, the Pi camera snaps an image of the individual and sends it through email to the remote user.

## 2.2. Remote Lock

The remote locking and unlocking capability allows users to manage door access when away from home through a mobile application. This eliminates the need to provide a spare key to short-term visitors, as the door can be locked and unlocked remotely. Additionally, the system enables the owner to add users remotely. Users can remotely unlock the door using a Telegram bot if the person is known to them [14]. Furthermore, users can control the door remotely through a Blynk mobile application on their smartphones. [15].

## 2.3. Notifications

There are two types of notifications: physical and virtual. Physical notifications involve using a loud alarm or flashing outdoor lights to alert the owners and neighbours to an intruder and deter them. Virtual notifications refer to emails or message alerts sent to the user when the system detects a stranger. Previous studies have explored similar approaches, such as using voice alerts or sirens to inform neighbours [14], sending emails with recorded images to the homeowner [12], and combining SMS with email notifications containing the intruder's image [16]. Additionally, Blynk also can be used to receive notifications of detected intruders [15].

## 2.4. Face Recognition

The process of facial recognition involves the identification or verification of an individual's identity through the analysis and comparison of their distinctive facial characteristics. This technology can be used to recognize people in images, videos, or in real-time. OpenCV, a free software library, is available for computer vision and machine learning. It can recognize objects, faces, and even handwriting in images and videos. The facial recognition process in a computer has three main steps: data gathering, training the recognizer, and recognition. During data gathering, facial information is collected for the individuals to be identified. The recognizer is then trained on this facial data and the corresponding names. Finally, the trained recognizer is used to identify people in new images [17].

Several IoT studies use OpenCV for system performance and accuracy, employing techniques such as Haar-Cascade for face detection and Eigenfaces [18], Fisherfaces [18], and Local Binary Pattern Histograms [3][18][19] algorithm for recognition. In a study focused on detecting human intruders, the researchers employed a two-step approach. First, they extracted distinctive visual features from the captured images using the Histogram of Oriented Gradients technique. Then, they performed a classification of these features by applying a Support Vector Machine algorithm to identify the presence of human intruders [13]. This low-cost solution significantly improves household security and suggests future enhancements for better accuracy and functionality.

The widespread use of face masks during the COVID-19 pandemic has posed a challenge to traditional facial recognition systems, making them inefficient [20]. LBPH has been utilised to detect occluded faces and tested on surgical mask users [3]. However, the result shows that it fails to detect occluded face users.

### 2.5. Two-Factor Authentication (2FA)

There are two types of authentication factors in 2FA. First is the knowledge factor, which refers to something the user knows, such as a password or PIN. Second is the possession factor, which is something the user has, such as an I.D. card and security token. Lastly, the inherence factor which refers to something inherent in the user's physical self, like fingerprint, voice, retina, and face [10].

A two-factor authentication door access control system utilizes both facial recognition and a one-time password [3]. The proposed system employs a two-factor authentication approach, whereby facial recognition serves as the first factor, and a one-time password (OTP) sent to the associated phone number constitutes the second factor. Access to the door is only granted upon successful verification of both the recognized facial features and the correct entered OTP.

Another system implements RFID and OTP as the two-factor authentication for door unlocking. First, the user scans a pre-assigned RFID key fob through an RFID reader. Then, the system recognizes the user, displays their name and a picture, and prompts them to enter a six-digit passcode. When the user enters the passcode, the system sends an OTP to the user's mobile device. The door is accessible once the user entered the correct OTP [4].

A lock system prototype has been constructed that initially verifies the user with RFID and then requires the user to enter the password sent by the system to gain access. The system uses the GSM module to send the specific code to the user. However, due to the nature of the GSM module, there is a delay in delivering generated code as a result of network failure or failure to send at all [5].

Another entryway prototype constructed fingerprint recognition and passcode identification to unlock the electromagnetic ZKteco LM2805 door lock [21]. Fingerprint recognition technology can be a reliable form of authentication, but its performance may be impacted by environmental influences or the physical condition of the user's fingerprints.

As indicated in Table 1, there appears to be a need to implement occluded face recognition using IoT without burdening the processor power while being able to accurately identify the personnel.

Table 1. Comparative review of pass entryway access control system

| Works | Motion Detection | Remote Control | Notification | Two Factor Authentication ||||||
|---|---|---|---|---|---|---|---|---|---|
| | | | | Knowledge | Possession || Inherence |||
| | | | | Password | OTP | RFID | Finger print | Face Recognition ||
| | | | | | | | | Full Face | Occluded |
| [3] | | | | | X | | | X | |
| [4] | | | | | X | X | | | |
| [5] | | | X | | X | X | | | |
| [12] | X | | X | | | | | | |
| [13] | X | | X | | | | | X | |
| [14] | X | X | X | | | | | | |
| [15] | X | X | X | | | | | | |
| [16] | X | | X | | | | | X | |
| [18] | | | | | | | | X | |
| [19] | | | | | | | | X | |
| [21] | | | | X | | | X | | |
| Proposed system | X | X | X | X | | | | X | X |

## 3. PROPOSED SYSTEM OVERVIEW

### 3.1. Prototype Design

This research proposed to improve research [3] by integrating motion detection, full face and occluded face recognition with password combination, remote control and virtual notifications. For this study, the system will be referred as Two-Factor Authentication Smart Entrway (2FASE).

The prototype design is centred around a Raspberry Pi 4 model B, a keypad, a 1602 LCD display, a PIR sensor module, a camera module, and a tiny buzzer (refer to Figure 1 and Figure 2). It incorporates a Pi Camera to capture live video frames for face detection and recognition. A PIR motion sensor is used to detect motion in front of the door, initiating the face recognition process. The LCD screen and keypad provide a user-friendly interface for input and system status display. Integration with Telegram is achieved through a Telegram bot, enabling real-time communication for access control and instant notifications to authorised users. When the PIR sensor detects a motion, face recognition will be activated to verify if that person's face exist in system or not. If a stranger is detected, it will use telegram to communicate with the owner. If the person's face is recognized, he or she will need to use the keypad to enter the password to unlock the door. The access is only granted when both face recognition and password verification is correct. Owner can lock and unlock the door as well as add user remotely using mobile application.

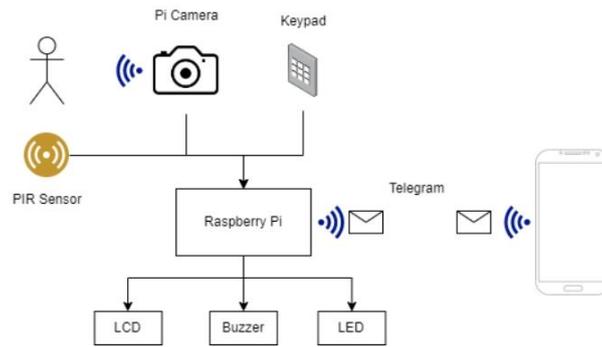

Figure 1. Prototype Design

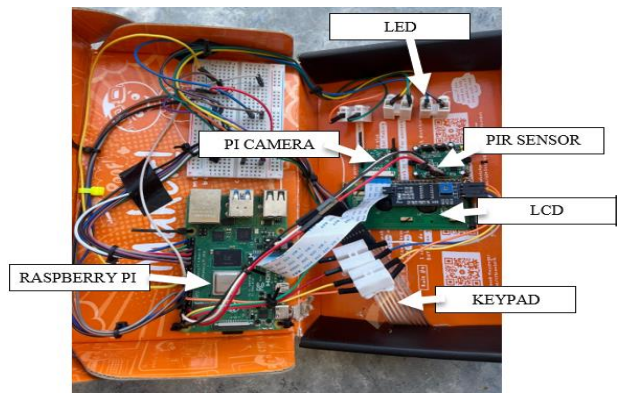

Figure 2. Prototype internal setup

## 3.2. Smart Door Integration with Face Recognition

The face recognition system is divided into two distinct modes. 'Mode 1' uses the LBPH algorithm to recognize full faces. The second mode employs a modified LBPH algorithm to identify occluded faces. The LBPH algorithm is known for its simplicity, efficiency, and ability to handle variations in lighting conditions. It works by dividing the face into small regions and comparing the patterns of pixels within each region. In Mode 1, the system utilizes a cascade classifier to detect and extract key facial features, such as the eyes and mouth, ensuring thorough coverage of the entire face (Figure 3). If the recognised face matches with the stored I.D., the system will grant access by unlocking the door. However, if the confidence level is below 70, indicating a potential mismatch, the system will flag the person as an unidentified individual. The system will show in the LCD message "face recognised," and then the photo will be saved as temp.jpg. And then the door will unlock.

If the confidence level is above 70%, the system will recognise the person as an unknown user. This functionality safeguards against unauthorized entry, preserving the system's security strength. The photo will then be taken and saved with the name stranger.jpg. The system will then prompt the administrator to confirm whether they recognize the face and ask if they would like to register the unknown individual as a new authorized user. However, if the admin does not recognise the face or chooses not to register them, the system will continue flagging the person as an unidentified individual.

In 'Mode 2', the LBPH (Local Binary Patterns Histogram) algorithm is modified to find and recognise partial faces (Figure 4). The results of 'Mode 2' are important in many ways. The fact that the system can be set up so that only part of the face is visible matches real-world situations where it might not be possible to see the whole face. For example, when users are wearing masks or coming at the camera from an angle, "Mode 2" uses the facial information that is available to to ensure the authentication process is accurate and reliable.

By dividing our system into two modes, each utilizing a specific approach, we aim to provide flexibility and cater to different scenarios. Mode 1 with the LBPH algorithm offers a reliable and efficient option for full face recognition, while Mode 2, with the modified LBPH recognizer, provides a solution for situations where only half face information is available.

The user can select the desired mode based on their specific requirements and priorities. This flexibility allows for customization according to factors such as the available computational resources, desired accuracy levels, and the specific application context.

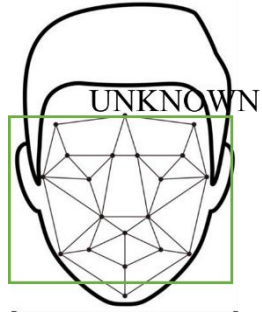 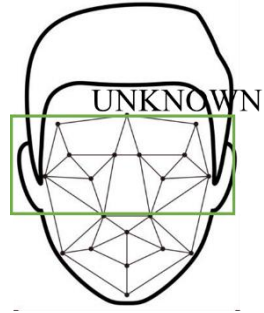

Figure 3 Full face recognition design       Figure 4 Occluded face recognition design

In the occluded face recognition, the system will determine the Bounding Box for the combined region of the eyes and nose:

Bounding Box = ($x_{eyes\_nose}$, $y_{eyes\_nose}$, $w_{eyes\_nose}$, $h_{eyes\_nose}$)   (1)
where,
$x_{eyes\_nose}$ = min ($x_{eye1}$, $x_{eye2}$)   (2)
$y_{eyes\_nose}$ = min ($y_{eye1}$, $y_{eye2}$, $y_{nose}$)   (3)
$w_{eyes\_nose}$ = max ($x_{eye1} + w_{eye1}$, $x_{eye2} + w_{eye2}$, $x_{nose} + w_{nose}$) - $x_{eyes\_nose}$   (4)
$h_{eyes\_nose}$ = max ($y_{eye1} + h_{eye1}$, $y_{eye2} + h_{eye2}$, $y_{nose} + h_{nose}$) - $y_{eyes\_nose}$   (5)

In cases where the nose is not visible, the system will solely scan the area around the eyes. This approach optimizes system resources and enhances accuracy by concentrating on specific facial features. Prioritizing the detection of eyes and the nose enables the system to offer a level of identification even when the complete face isn't visible.

### 3.3. Telegram Bot and Notifications

The prototype is integrated with the telegram bot. Admin can use these commands, which are "capture", "unlock", "lock", "change_", "mode1", "mode2", "adduser_", and "showpassword".

### 4. TESTING AND RESULTS

The test design for full face recognition would involve capturing various angles and expressions of the user's face to ensure accurate identification in different scenarios. Additionally, it would include testing the system's ability to handle changes in lighting conditions or facial hair. For partial face recognition, the test design would focus on evaluating its speed and reliability in quickly authenticating users based on partial facial features while also testing its resistance to spoofing attempts.

In this study, 4 users were used to collect 150 images datasets each for various tests, including users with head cover and different expressions.

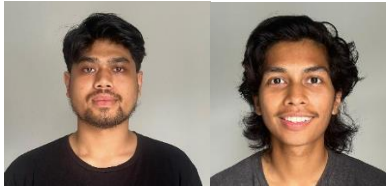
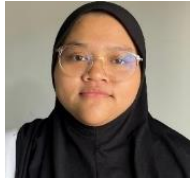
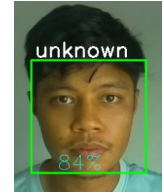

Figure 5. Non-head-covered Registered User

Figure 6. Head-covered Registered User

Figure 7. Unregistered User

### 4.1. Face Recognition Accuracy Test

The objective of the face recognition test is to determine the model's accuracy in recognising and granting access to enrolled users when a full face or partial face is detected. As depicted in Table 2, the model demonstrates a 51% recognition confidence level for enrolled users without head coverings. In contrast, the head-covered registered users are recognised 45% of the time if they use a full face model. As for the occluded face model, the result indicates a reduced confidence level where non-head-covered registered users can be recognised at 48.38% while the head-covered registered users are at 55.2% (Fig. 8 – 9). This shows that the full face model works best on the head-covered user, while the partial face model works best with the non-head-covered user.

Table 2 Result of Face Recognition Model

| Registered User | Full face Model (%) | Occluded face Model (%) |
|---|---|---|
| Wanhariz | 51 | 48.38 |
| Najwa | 45 | 55.20 |
| Nazrin | 34 | 39.80 |

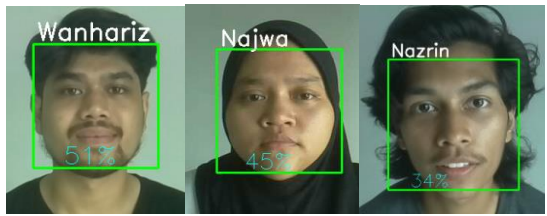
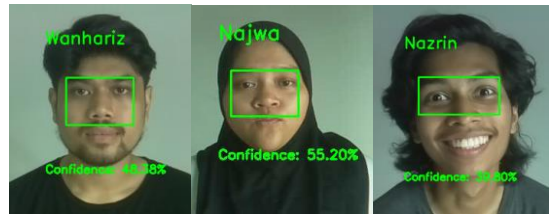

Figure 8. Full face model

Figure 9. Occluded face model

### 4.2. Face Expression Recognition Test

The objective of the face recognition test is to determine the model's accuracy in recognising users with normal, happy and sad expressions using both models. It is important to consider that lower percentages indicate higher accuracy in this context. Table 3 presents the results of face expression recognition test. Overall, the system is able to recognize the user in all expression. The analysis reveals varying performance across users and expressions for both models, with a notable trend of reduced accuracy (higher percentages) for the "Sad" expression compared to "Normal" and "Happy." The Full Face Model generally performs better for Najwa, particularly for "Happy" and "Sad," while the Occluded Face Model achieves better accuracy for the "Sad" expression in Wanhariz and slightly outperforms for "Normal" and "Happy" in Nazrin. However, the Occluded Face Model shows less consistency across users and expressions, struggling particularly with Najwa's "Sad" expression. User-specific variability is evident, as Wanhariz and Najwa display larger performance gaps between expressions, while Nazrin's accuracy is more stable but lower overall. These trends highlight the need for enhanced robustness in recognizing the "Sad" expression and improved handling of occlusions to achieve consistent results across all users and expressions.

In another aspect, the occluded face model outperforms full face recognition for non-head-covered users, whereas the full face model outperforms occluded face recognition for the head-covered user for either Normal or Sad or Happy expressions.

Table 3. Result of Face Expression Recognition Model

| Registered User | Expression | Full face Model (%) | Occluded face Model (%) |
|---|---|---|---|
| Wanhariz | Normal | 59 | 56 |
|  | Happy | 58 | 54 |
|  | Sad | 55 | 43 |
| Najwa | Normal | 44 | 55 |
|  | Happy | 42 | 51 |
|  | Sad | 53 | 56 |
| Nazrin | Normal | 41 | 38 |
|  | Happy | 42 | 41 |
|  | Sad | 44 | 52 |

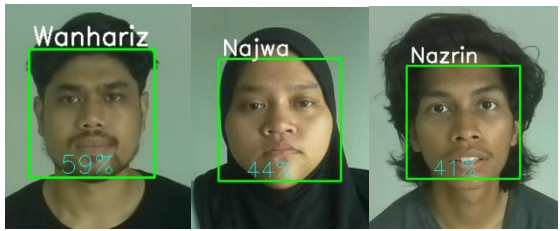

Figure 10. Full face recognition with a normal expression

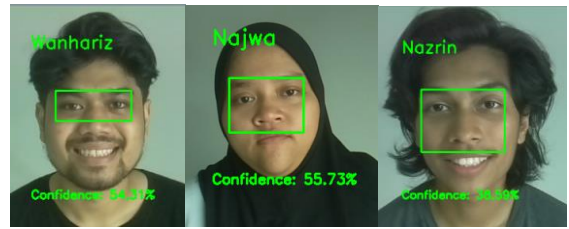

Figure 11. Occluded face recognition with a normal expression

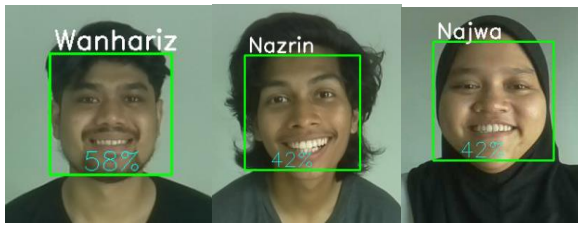

Figure 12. Full face recognition with a happy expression

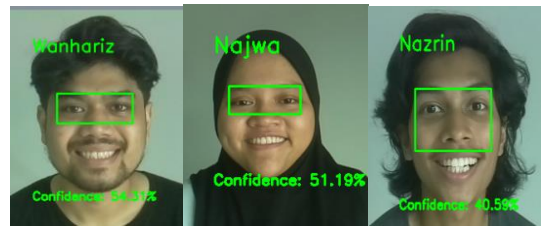

Figure 13. Occluded face recognition with a happy expression

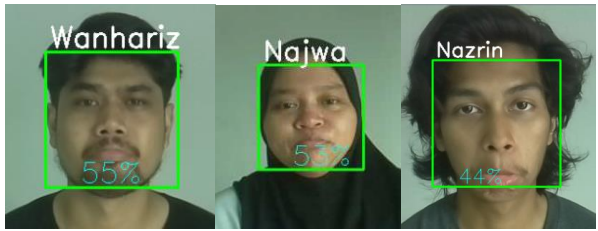 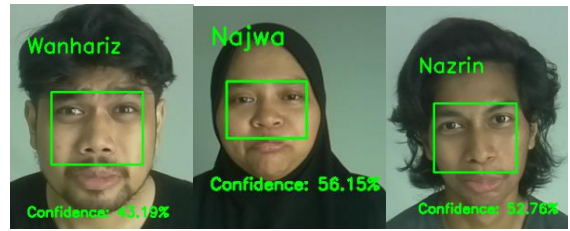

Figure 14. Full face recognition with a sad expression

Figure 15. Occluded face recognition with a sad expression

### 4.3. Masked Face Recognition Test

In this experiment, we assessed the system's capability to accurately recognise users even when they are wearing masks.

Table 4. Result Of Masked On Using Occluded Face Recognition Model

| User | Recognition Confidence (%) | Result |
|---|---|---|
| Nazrin | 59 | Pass |
| Wanhariz | 52 | Pass |
| Najwa | 68 | Pass |

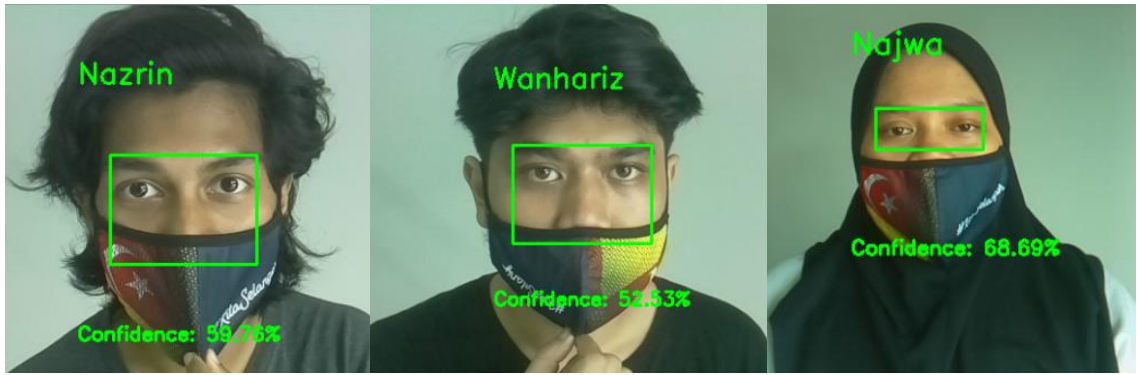

Figure 16. Masked Face Registered user using Occluded Face Recognition

Based on Table 4, the result shows that Wanhariz's face recognition was the most accurate, with the system showing the highest confidence in the match. Nazrin's results were also quite good, while Najwa's higher percentage of 68% indicates relatively lower recognition accuracy, though still within acceptable parameters as the system can recognize these users. The variation in scores might be influenced by how the mask fits each user's face. Wanhariz's better score could indicate their mask was positioned in a way that maintained key facial feature visibility. The system appears to perform better when distinctive upper facial features are present. The eyes and nose areas become the crucial recognition points when masks are worn. In Figure 16, the bounding box include the eyes and nose areas, thus produced superior accuracy. Although user Nazrin's bounding box includes eyes and nose areas, the masks has covered the lower region of the nose which reduces the confidence of the recognition rate. Lastly, because user Najwa has covered her nose area with mask, the recognition rate is less accurate as the bounding box is on the eye area. All users passed despite varying scores, suggesting the system has a robust tolerance range for masked face recognition.

### 4.4. Unregistered User Face Recognition Test

The system shows unregistered user face recognition result between the two face models (Refer Figure 17a-b). The Occluded Face Model yield a slightly higher rate compared to Full Face Model. Consistent with the system's design, unrecognized faces are classified with a confidence rate exceeding 70%.

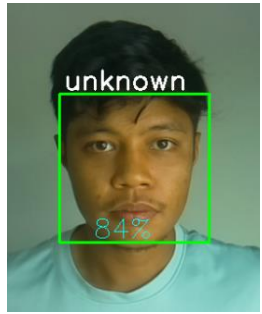
Figure 17a. Unregistered user using full face recognition

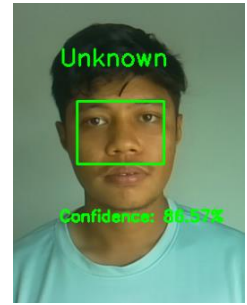
Figure 17b. Unregistered user using occluded face recognition

### 4.5. Face Detection Performance Analysis

Table 5. Confusion Metrics Comparison of Face Recognition Models

| Model | Evaluation Parameters | | |
| --- | --- | --- | --- |
| | *Accuracy* | *Precision* | *Recall* |
| Full Face | 80.83% | 91.48% | 87.77% |
| Occluded Face | 70% | 80% | 83.26% |

Table 5 shows the overall performance of the face recognition system, the average values of Accuracy, Precision, and Recall provide a comprehensive assessment for all tested user. The system demonstrated an overall accuracy of approximately 80.83%, a precision of approximately 91.48%, and a recall of approximately 87.77% across the tested users, indicating its general reliability in recognizing known users and differentiating them from unrecognized individuals. The findings establish a robust groundwork for the system's effectiveness. The high precision rates imply a low incidence of false positives, while the elevated recall values indicate the system's proficiency in accurately recognizing users across diverse scenarios. While the Occluded Face model achieved an Accuracy of approximately 70%, a Precision of approximately 80%, and a Recall of approximately 83.26% across all tested users. These averages indicate the system's general reliability in recognizing known users. However, there is room for enhancement to further increase precision and recall, especially when dealing with unrecognized individuals. Overall, these results serve as a valuable foundation for system refinement and optimization to ensure its effectiveness in real-world scenarios.

Table 6. Result of Face Recognition Test on Different Distance

| Distance (cm) | Result |
| --- | --- |
| 40 | Correct |
| 50 | Correct |
| 60 | Correct |
| 70 | Correct |
| 80 | False |
| 90 | False |

Based on Table 6, the maximum distance that the Pi camera could recognized the user is 70cm. This limitation may be attributed to the reliance on the bounding box area to detect the users' eyes and nose, which could be impacted by factors such as distance and lighting conditions.

Based on the comprehensive testing of the smart door lock system using both full face recognition and partial face recognition, the results indicate notable differences in recognition speed for different users and recognition types. In the case of full face recognition, users experienced varying recognition times, ranging from as low as 35 seconds to as high as 41 seconds. The success rates also exhibited variability among users and attempts. On the other hand, occluded face recognition generally demonstrated faster recognition times, with users experiencing speeds as low as 13 seconds to 15 seconds. Despite instances of recognition failure, occluded face recognition consistently exhibited quicker response times compared to full face recognition. The selection between the recognition methods should consider not only speed, but also accuracy and overall system performance. These factors collectively influence the selection of the most suitable recognition approach for the smart door lock system.

### 4.6. Password Authentication Test

The system is configured to automatically transition back to the facial recognition phase if the user does not successfully enter the password within the allotted 30-second time period, as shown in Figure 18a. Figure 18b shows that if a password is entered incorrectly more than three times, then go back to the facial recognition phase. Figure 18c respectively shows that candidates Nazrin have successfully pass face recognition test. Table 7 shows the results of password verification based on different situation.

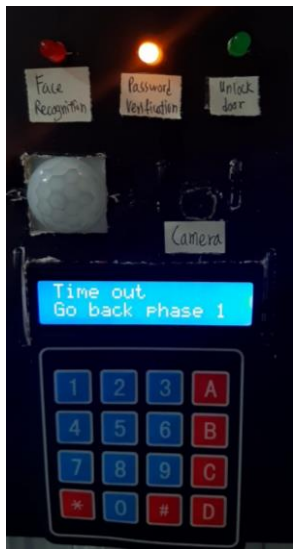
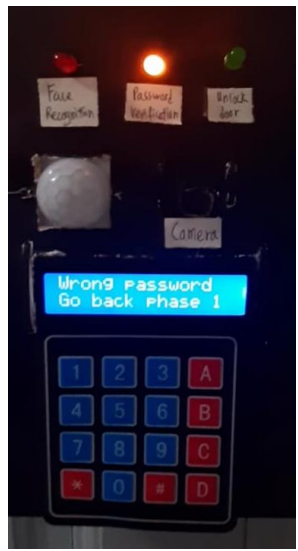
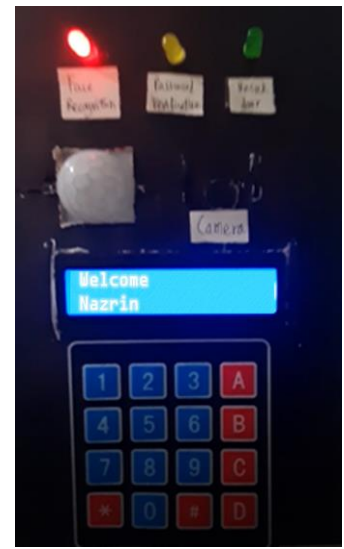

Figure 18a. Time out for password verification (30 seconds)

Figure 18b. Password entered wrong more than 3 times

Figure 18c. Candidates Nazrin passed face recognition phase

Table 7 Result of password verification based on different situation

| No | User | Actual password | Input password | No of tries | Unlock door |
|---|---|---|---|---|---|
| 1 | Nazrin | 7816 | 7514 | 1 | ✕ |
|   |   |   | 6447 | 2 | ✕ |
|   |   |   | 1489 | 3 | ✕ |
| 2 | Nazrin | 7816 | 4718 | 1 | ✕ |
|   |   |   | 7416 | 2 | ✕ |
|   |   |   | 7816 | 3 | ✓ |
| 3 | Nazrin | 7816 | 7816 | 1 | ✓ |

### 4.7. Smart Door Lock and Telegram Bot Integration Test

During the integration test, the smart door lock is connected to a Telegram Bot, allowing admin to control the lock remotely through the telegram app. When the PIR sensor detect a presence, and the face detection did not recognize the user, it will notify the admin through the Telegram (Figure 19). User registration is successful by using the command adduser_Nazrin. The system then proceeded to capture Nazrin's dataset by analyzing various face expressions. The integration test is successful as admin is able to lock and unlock the door using commands sent via Telegram. And the notification is successfully received .

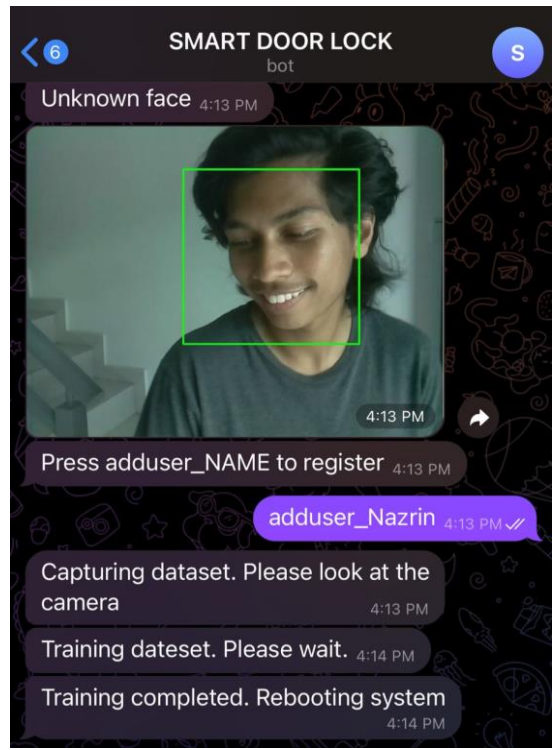

Figure 19. Unknown user notification in Telegram

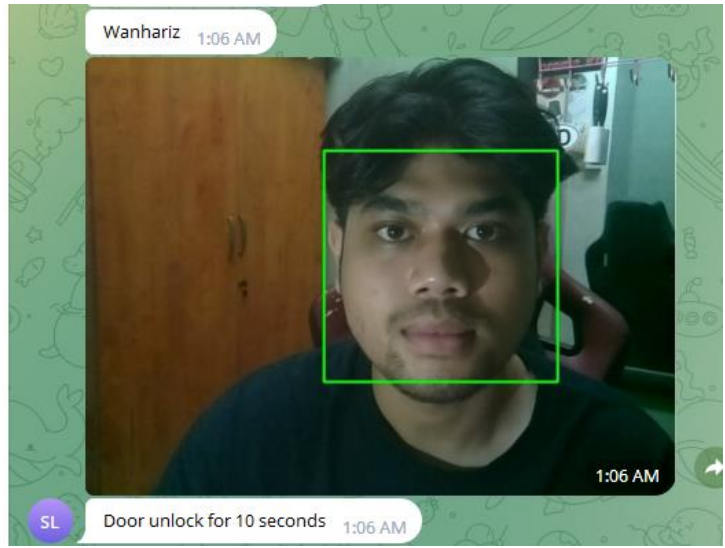

Figure 20. Notification alert of user unlocking the door.

## 4.8. User Acceptance Test (UAT)

A usability assessment was undertaken to examine the influence of the system's design on users' interactions with the Smart Entryway Access Control (2FASE) system. A preliminary study using a quantitative methodology is surveyed. A structured questionnaire was used to collect data from 20 individuals with different levels of I.T. skills. Technology Acceptance Model (TAM) by [22] and [23] will be adopted for this study. The model encompasses four central factors for evaluation: *Perceived Usefulness*, *Perceived Ease of Use*, *Trust*, and *Behavioural Intention to Use*[22].

As shown in Table 8, the mean value was 4.45, indicating that the respondents generally agreed on the perceived usefulness of the 2FASE system evaluated in this study. The average mean value in Table 9 was 4.4, showing that, overall, the respondents agreed with the ease of use of the 2FASE system in this study. Based on Table 10, the respondents have a high level of trust in 2FASE, as indicated by the average mean result of 4.33, which is above 4. The average mean value in Table 11 is 4.17, which shows that the respondents agreed on their intention to use the system.

Table 8 Analysis for Perceived Usefulness towards System

| Analysis Category | Mean | St Dev | Avg Mean |
|---|---|---|---|
| **Perceived Usefulness (P.U.)** | | | |
| PU1. Using the 2FASE would enable me to unlock the door more quickly. | 4.6 | 0.50 | 4.45 |
| PU2. Using 2FASE would make it easier for me to unlock the door. | 4.45 | 0.60 | |
| PU3. Using 2FASE can notify me of any unusual activities that occur in front of the door. | 4.65 | 0.49 | |
| PU4. Using 2FASE would significantly increase the quality or output of my life. | 4.1 | 0.31 | |

Table 9. Analysis for Perceived Ease of Use Towards System

| Analysis Category | Mean | St Dev | Avg Mean |
|---|---|---|---|
| **Perceived ease of use (PEOU)** | | | |
| PEOU1. Learning to use 2FASE is easy for me. | 4.45 | 0.76 | 4.4 |
| PEOU2. I find my interaction with the 2FASE is clear and understandable. | 4.4 | 0.50 | |
| PEOU3. I think using 2FASE is easy. | 4.35 | 0.59 | |

Table 10. Analysis of Trust towards System

| Analysis Category | Mean | St Dev | Avg Mean |
|---|---|---|---|
| **Trust (T.R.)** | | | |
| TR1. 2FASE is trustworthy. | 4.25 | 0.44 | 4.33 |
| TR2. 2FASE provides reliable information. | 4.4 | 0.50 | |

Table 11. Analysis of Behavioural Intention to Use System

| Analysis Category | Mean | St Dev | Avg Mean |
|---|---|---|---|
| **Behavioural intention to use (B.I.)** | | | |
| BI1. I am willing to use 2FASE in the near future. | 4.25 | 0.64 | 4.17 |
| BI2. I will recommend 2FASE to others. | 4.3 | 0.47 | |
| BI3. I will continue using 2FASE in the future. | 3.95 | 0.60 | |

## 5. CONCLUSIONS

This study proposed a Two-Factor Authentication Smart Entryway Using Modified LBPH Algorithm. By combining full and occluded face recognition with real-time remote access control through a Telegram Bot, the prototype makes a big difference in the fields of access control and biometric recognition systems. This integration promotes enhanced security and user-centric design, offering greater versatility and usability compared to conventional access control systems. Through the UAT, users were more inclined to use the design of 2FASE in their residences. Also, the system can accurately identify users even when wearing masks or partial face coverings. It meets a modern need and is helpful in today's health and safety concerns. However, the accuracy is limited by low light or users' heavy disguises. Although 2FASE system involves complex recognition algorithms, the potential for false positives and false negatives remains an important issue. For future research, the feasibility of implementing alternative biometric authentication methods like iris or fingerprint recognition on IoT devices should be explored. Additionally, the accuracy of facial recognition systems can be enhanced through advanced algorithms and machine learning approaches. Furthermore, it would be beneficial for future work to address the ethical concerns surrounding facial recognition systems, such as privacy invasion and potential bias [24].

## ACKNOWLEDGEMENTS

The authors would like to thank the Center for Advanced Computing Technology (C-ACT), Fakulti Teknologi Maklumat dan Komunikasi, Universiti Teknikal Malaysia Melaka (UTeM) for supporting financially the work done in this paper.

## AUTHORS

**Zakiah Ayop** graduated with Bachelor of Computer Science and Master of Science in Distributed Computing from Universiti Teknologi Malaysia and Universiti Putra Malaysia respectively. Currently, she is a senior lecturer in Department of Computer System and Communications, Faculty of Information and Communication Technology (FTMK), UTeM. Her research interest is Information System, Internet of Things (IoT), Cyber-physical System (CPS) and Networking. She is a certified CCNA Instructor. Her work has been presented at various international conferences and published in journals, contributing significantly to advancements in her fields of interest.

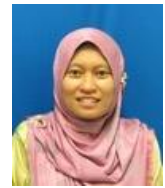

**Syarulnaziah Anawar** holds her Bachelor of Information Technology from UUM, Msc in Computer Science from UPM, and PhD in Computer Science from UiTM, Malaysia. She is currently a Senior Lecturer at the Department of Computer and Communication System, Faculty of Information and Communication Technology, UTeM. She is a member of the Information Security, Digital Forensic, and Computer Networking (INSFORNET) research group. Her research interests include human-centered computing, participatory sensing, mobile health, usable security and privacy, and societal impact of IoT. You can contact her at email syarulnaziah@utem.edu.my

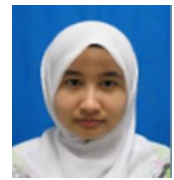

**Nur Fadzilah Othman** received a degree in Computer Engineering in 2008 and master in Educational Technology in 2011 at Universiti Teknologi Malaysia (UTM). In 2017, she obtained her PhD in the field of Information Security at Universiti Teknikal Malaysia Melaka (UTeM). She started his career as a senior lecturer at the Department of Computer System and Communication, Faculty of Information Technology and Communication, UTeM from March 2018. She is an active researcher and has been written and presented a number of papers in conferences and journals.

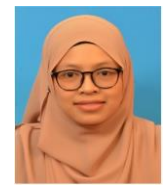

**Wan Mohamad Hariz Bin Wan Mohamad Rosdi** holds his Diploma in Computer Science from KPM Beranang, and Bachelor of Computer Science in Cybersecurity from UTeM, Malaysia. He is currently a Penetration Tester at AskPentest. His research interests include cloud security, IoT security and automation for penetration testing. You can contact him at email wnharz@gmail.com

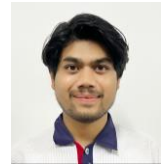

**Looi Wei Hua** holds his Bachelor of Computer Science in Computer Networking from UTeM, Malaysia. He is currently a Network Engineer at V6 Technology. His research interests include cloud security, IoT security and automation for network security.

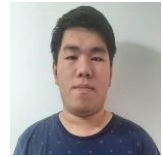